\documentclass[11pt]{article}

\usepackage{acl}

\usepackage{xcolor}
\usepackage{times}
\usepackage{latexsym}
\usepackage[T1]{fontenc}
\usepackage[utf8]{inputenc}
\usepackage{microtype}
\usepackage{inconsolata}
\usepackage{graphicx}
\usepackage{multirow}
\usepackage{booktabs}
\usepackage{subcaption}
\usepackage{algorithm}
\usepackage{algpseudocode}
\usepackage{pifont}
\usepackage{amsmath}

\newcommand{\cmark}{\color[HTML]{036400}{\ding{51}}}
\newcommand{\xmark}{\color[HTML]{9A0000}{\ding{55}}}

\usepackage{enumitem}

\title{``Whose Side Are You On?'' \\ Estimating Ideology of Political and News Content Using Large Language Models and Few-shot Demonstration Selection}

\author{
Muhammad Haroon
\\
{\normalsize University of California, Davis}
\\
{\normalsize \texttt{mharoon@ucdavis.edu}}
\And
Magdalena Wojcieszak
\\ 
{\normalsize University of California, Davis}
\\
{\normalsize \texttt{mwojcieszak@ucdavis.edu}}
\And
Anshuman Chhabra
\\
{\normalsize University of South Florida}
\\
{\normalsize \texttt{anshumanc@usf.edu}}
}

\begin{document}
\maketitle
\begin{abstract}
The rapid growth of social media platforms has led to concerns about radicalization, filter bubbles, and content bias. Existing approaches to classifying ideology are limited in that they require extensive human effort, the labeling of large datasets, and are not able to adapt to evolving ideological contexts. This paper explores the potential of Large Language Models (LLMs) for classifying the political ideology of online content through in-context learning (ICL). Our extensive experiments involving demonstration selection in label-balanced fashion, conducted on three datasets comprising news articles and YouTube videos, reveal that our approach outperforms zero-shot and traditional supervised methods. Additionally, we evaluate the influence of metadata (e.g., content source and descriptions) on ideological classification and discuss its implications. Finally, we show how providing the source for political and non-political content influences the LLM's classification.
\end{abstract}

\section{Introduction}
Social media platforms are one of the leading sources of information for the majority of Americans and have grown in centrality over the past few years \cite{pew2020measuringnews}. 
Observers worry that the content recommended and seen on social media platforms can radicalize users, cause extremism \cite{tufekci2018youtube, roose2019makingytradicalNYT}, and facilitate the formation of filter bubbles that restrict exposure to diverse viewpoints \cite{pariser2011filter}. 

\looseness-1 However, empirical evidence for these worries is inconsistent \cite{haroon2023auditing,hosseinmardi2021examining, chen2023subscriptions, nyhan2023like} in part because scholars use different methodologies and approaches (audits versus observational data) and in part because we lack \textit{robust} (i.e., highly accurate across domains) and scalable (i.e., requiring minimal human supervision) methods for estimating political ideology of the content users consume.\footnote{We recognize the need to also estimate political ideology of the users themselves, which has been done using variety of methods \cite{barbera2015birds} and which is outside the scope of this work.} Systematic assessments of whether (and the extent to which) social media platforms indeed expose users to ideologically biased or politically radical content are not possible without \textit{robust} and \textit{scalable} political ideology estimation methods. 

\looseness-1 Prior approaches to ideology estimation have utilized manual expert annotations \cite{otero2019ad}, supervised classification models \cite{baly2020detect}, and social media sharing patterns \cite{le2017scalable,barbera2015birds,lai2024estimating,wojcieszak2023polarized,eady2020news}. Manual annotations are highly accurate but resource-intensive and difficult to scale, especially given the growing volume of online content. Supervised classification models, while offering some automation, face challenges such as a reliance on extensive labeled data and difficulty adapting to dynamic ideological contexts. Social media-based methods estimate ideology by analyzing content sharing patterns on platforms such as Twitter or Reddit, but these approaches are hindered by data access limitations \cite{murtfeldt2024riptwitterapieulogy}, reliance on widely shared content, and being agnostic to the content itself. These limitations highlight the need for scalable, adaptable, and robust solutions to ideology estimation.

\looseness-1 
In this paper, we advance on literature showing that Large Language Models (LLMs) can be applied to ideology estimation of text in the context of the two-party US political spectrum ranging from democrat/liberal to republican/conservative, and can address the aforementioned challenges of scalability and robustness. More specifically, pre-trained LLMs can learn new classification tasks \textit{in-context} at test-time \cite{hardttest}, which enables a few-shot learning paradigm that only requires a few annotated samples. Furthermore, LLMs can handle complex language subtleties necessary for accurate/robust ideology estimation. They can also adapt to new and evolving information by simple adjustments of the provided prompt. Thus, LLMs can serve as a compelling choice for robust and scalable ideology estimation across diverse text domains and platforms.

\looseness-1 Past applications of LLMs in ideology estimation \cite{heseltine2024llm,tornberg2023chatgpt4outperformsexpertscrowd,gilardi2023chatgpt} have primarily focused on zero-shot approaches, utilizing pre-trained LLMs without employing more nuanced few-shot in-context learning (ICL) strategies. Our work advances these prior studies by proposing a modified version of coverage-based ICL that selects demonstrations in a class-balanced manner, substantially improving performance over simple zero-shot prompting and coverage-based ICL. Moreover, whereas existing works primarily classify political texts with explicit ideology (e.g., tweets from senators), we demonstrate the effectiveness of our approach across diverse content categories, including political news, political non-news, and non-political news content.

In sum, we make the following contributions:
\begin{itemize}[nosep]
    \item We conduct an extensive study on the capabilities of differently sized LLMs (Llama2-13B, Mistral-7B, and GPT-4o) in estimating the ideology of text. Through rigorous experimentation on three distinct datasets previously labeled by domain experts, we demonstrate these models’ accuracy, adaptability, and robustness in capturing ideological cues across multiple data. Our findings position LLMs as a robust and scalable solution for effective ideology estimation.
    \item We propose a demonstration selection algorithm that utilizes coverage-based ICL and class-balanced selection to achieve state-of-the-art LLM performance on ideology estimation. We show that our approach improves upon zero-shot prompting used in past work \cite{heseltine2024llm,tornberg2023chatgpt4outperformsexpertscrowd,gilardi2023chatgpt}.
    \item We analyze the effect of additional metadata, such as content source and description, on LLM ideology predictions, especially for neutral or non-political content. Our results show that metadata can influence model predictions, at times aligning the predicted ideology with that of the source or introducing bias when source and content ideologies diverge. This analysis provides critical insights into the limitations and potentials of LLMs for nuanced ideological analysis.
\end{itemize}

\section{Related Work}
Prior approaches to ideology estimation have generally fallen under three types (1) manual expert annotations \cite{otero2019ad}, (2) content analysis through supervised classification models \cite{baly2020detect}, and (3) social media sharing approaches \cite{le2017scalable,barbera2015birds,lai2024estimating,wojcieszak2023polarized,eady2020news}. 

\noindent \textbf{Manual annotations by experts.} Several efforts have been taken by independent parties to identify the political ideology of online news articles and sources such as Ad Fontes \cite{otero2019ad}, AllSides\footnote{\url{https://allsides.com}}, Media Bias/Fact Check\footnote{\url{https://mediabiasfactcheck.com}}, etc through a manual annotation process involving experts.
Whereas manual annotation is accurate and robust (albeit depending on the number and quality of experts employed), it is highly resource-intensive and not scalable. Expert annotation becomes especially challenging as the volume of social media content increases consistently. 

\noindent \textbf{Training supervised classifiers.}
In turn, supervised ideology classification models can provide some form of automation. The advent of deep learning techniques revolutionized this field by enabling more sophisticated models capable of capturing complex linguistic patterns. \citet{iyyer2014political} introduced deep neural networks, such as LSTMs and CNNs, which outperformed traditional models in detecting ideology from unstructured text data. In a similar vein, \citet{baly2020detect} used a BERT model trained on political news articles to show improvements over existing state-of-the-art supervised learning approaches. 
However, they are limited by a need for large volumes of human-labeled data and have difficulty adapting to dynamic and ever evolving ideological contexts (e.g., cross-platform estimation, emerging news events, changes in the political environment). 

\noindent \textbf{Utilizing social media data.}
Social media sharing ideology estimation approaches utilize data from external platforms such as Twitter \cite{le2017scalable,barbera2015birds,wojcieszak2023polarized,eady2020news} or Reddit \cite{lai2024estimating} to assess how a given piece of content is shared among users. By identifying left- and right-leaning elites (e.g., politicians, news media) those users follow, those approaches first estimate the ideology of the users, and then estimate the ideology of the shared content by aggregating the ideology of the sharing users. This approach is limited in several ways. First, as it relies on third-party platforms (such as Twitter/X), changes to APIs and data acquisition systems can render this approach untenable. For instance, since Twitter transitioned to X in 2023, it has become challenging for researchers to collect sharing data at scale without incurring significant costs \cite{murtfeldt2024riptwitterapieulogy,poudel2024postapi}. Second, to accurately estimate the ideology of content (e.g., a YouTube video), the content needs to be shared extensively, which might not happen for niche videos with small viewership. Finally, this approach is agnostic to the nature of the content itself and only looks at how it is shared by users. Owing to these limitations, these approaches may not be robust and reliable.

\noindent \textbf{Using Large Language Models.}
LLMs have achieved state-of-the-art performance for several tasks across multiple language domains (e.g., legal transcripts \cite{colombo2024saullm} and medical data \cite{yang2022large}) and contexts (e.g., identifying misinformation \cite{hoes2023using}, hate speech \cite{huang2023chatgpt}, sentiment \cite{heseltine2024llm}, or relevance, topics, and frames in tweets and news articles \cite{gilardi2023chatgpt}. 

As aforementioned, prior works have utilized LLMs to classify the ideology of text \cite{heseltine2024llm,tornberg2023chatgpt4outperformsexpertscrowd,gilardi2023chatgpt}. As we show in Table \ref{tab:related-comparison}, our work differs from these in several key ways. First, all these approaches are zero-shot methods that only utilize the pre-trained LLMs as is, and do not employ optimal and efficient few-shot in-context learning (ICL) approaches to improve accuracy and adapt LLMs to multiple domains. We propose a modified version of coverage-based ICL that selects demonstrations in a class-balanced manner, and improves performance over simple zero-shot prompting and coverage-based ICL. Second, these methods focus on classifying texts from sources with clearly distinct ideology (e.g. tweets from liberal or conservative politicians). We focus on content with differential levels of ideological signal to demonstrate that our LLM-based approach can estimate ideology of \textit{political news}, \textit{political content but not news}, and \textit{non-political news content} as well. We also analyze the potential limitations for ideology estimation of each of these content subcategories. Finally, prior works do not undertake extensive benchmarking of performance across multiple LLMs and diverse datasets. We run experiments on three different datasets (two of news articles and one consisting of YouTube videos) and three different LLMs (open-source and closed-source) of different sizes. Our results shed light on which LLMs are better suited for ideology classification and what data/information augments LLM performance, i.e. whether additional video descriptions or channel information in prompts increases/reduces performance.

\begin{table}[t]
\centering
\vspace{-2mm}
\resizebox{0.47\textwidth}{!}{%
\begin{tabular}{lccc}
\toprule
\multirow{2}{*}{\textbf{Method}} & \multirow{2}{*}{\textbf{Zero-shot}} & \textbf{Few-shot} & \textbf{Non-political}\\
 & & \textbf{(ICL)} & \textbf{content} \\
\midrule
\citet{gilardi2023chatgpt} & \cmark & \xmark& \xmark \\
\citet{tornberg2023chatgpt4outperformsexpertscrowd} & \cmark & \xmark  & \xmark \\
\citet{heseltine2024llm} & \cmark &  \xmark & \xmark \\
\midrule
\textbf{Ours} & \cmark & \cmark & \cmark \\
\bottomrule
\end{tabular}
}
\caption{Comparison of our work with prior LLM-based approaches.}
\label{tab:related-comparison}
\vspace{-3mm}
\end{table}

\section{Problem Formulation}

Our research problem can be defined as follows: \textit{Given content that has an underlying ideological leaning (e.g., a YouTube video, a news article), we aim to determine if the content is liberal, neutral, or conservative.}
Next, we describe our proposed approach where we utilize set coverage and balanced shot selection for ideology classification. 
%




\subsection{Proposed Approach}


Algorithm \ref{alg:ideology_estimation} provides our proposed in-context learning (ICL) framework.
Given a LLM $\mathcal{L}$, a training set of content with ideology class labels (Liberal, Neutral, Conservative) $\mathcal{T}$, and number of demonstrations $k$, our task is to classify a given query $x$ (an online post, video, or news article) as either liberal, neutral, or conservative.
Our approach first builds a smaller candidate set $\mathcal{C}$ of potential demonstrations and computes a relevance score for each demonstration in the candidate set.
From the sorted candidate set, we select $k$ demonstrations equally from across all three ideology classes, i.e., only $\frac{k}{3}$ demonstrations can be selected from each ideology class.
Thus, we skip over demonstrations with higher scores if $\frac{k}{3}$ demonstrations have already been selected from the class they belong to.
The instruction prompts used can be found in Appendix \ref{appendix:task_prompts}.

\begin{algorithm}[t]
\fontsize{9}{9}\selectfont
\begin{algorithmic}[1]
  
  \State \textbf{Input:} 
  \begin{itemize}
      \item Test Instance: $x$
      \item LLM: $\mathcal{L}$
      \item Prompt: $\mathcal{P}$
      \item Training Set: $\mathcal{T}$
      \item Number of Demonstrations: $k$
  \end{itemize}
  \State \textbf{Output:} Ideology classification for $x$
  \State \textbf{Initialize:} 
  \begin{itemize}
      \item $\mathcal{D} \gets \emptyset$ \Comment{{\small Selected Demonstrations}}
      \item $\mathcal{C} \gets$ \textsc{Set}($\mathcal{T}$) \Comment{{\small Candidate set using coverage}}
      \item $counts(s) \gets 0$ $\forall s \in ideologies$ \Comment{{\small Label counts}}
  \end{itemize}
  
  \State \textbf{Sort} $\mathcal{C}$ (descending) by \textsc{Set-BSR} over $x$
  \State $i \gets 0$

  \While{$\sum counts \leq n$}
      \State $s \gets ideology(\mathcal{C}_{i})$
      \State
      \If{$counts(s) \leq \frac{n}{3}$} \Comment{{\small Ensure label balance}}
        \State $\mathcal{D} \gets \mathcal{D} \cup \mathcal{C}_{i}$ \Comment{{\small Add demonstration to $\mathcal{D}$}}
        \State $counts(s) \gets counts(s) + 1$ 
      \EndIf
      \State $i \gets i + 1$ \Comment{{\small Move to next element in $\mathcal{C}$}}
  \EndWhile
  
  \State \Return $\mathcal{L}(\mathcal{P} \cup \mathcal{D}, x)$ \Comment{{\small Classify $x$ with prompt $\mathcal{P}$ and selected demonstrations $\mathcal{D}$}}

\end{algorithmic}
\caption{Estimating Ideology using Set Cover and Label-Balanced Shot Selection}
\label{alg:ideology_estimation}
\end{algorithm}

\subsection{Test Query Specifications}
We specify the query $x$ to be classified as an online post, video, or news article that has a title, source, and a description.
An LLM can leverage information from all three components to aid in classification, however, limitations such as context-window sizes and longer compute times for longer text make it worthwhile to explore the minimum amount of information that provides the best accuracy.
As such, our experiments rely on four combinations: title; title and source; title and description; and title, source, and description.

\begin{table}[t]
\resizebox{.99\linewidth}{!}{
\setlength\fboxsep{1.5pt}
\begin{tabular}{p{0.5\textwidth}}
\toprule
\textbf{Sample Text} \\
\colorbox{blue!30}{Trump} \colorbox{red!30}{tweets} \colorbox{green!30}{COVID-19} \colorbox{brown!30}{video} so \colorbox{magenta!30}{outrageous} it was \colorbox{orange!30}{banned} from \colorbox{yellow!30}{Facebook} and \colorbox{purple!30}{Twitter} \\
\midrule
\textbf{Selected Demonstrations} \\
\vspace{-2.5mm}
\colorbox{purple!30}{Twitter} removes \colorbox{blue!30}{Trump} retweet that falsely claimed \colorbox{green!30}{COVID-19} ‘has a cure’ \\
\vspace{-2.5mm}
Reporter to sue Michigan Gov. Whitmer for allegedly not releasing \colorbox{green!30}{COVID-19} nursing home data \\
\vspace{-2.5mm}
AU Health masters \colorbox{green!30}{COVID-19} therapy but loses \$20 million \\
\vspace{-2.5mm}
\colorbox{yellow!30}{Facebook} to label \colorbox{green!30}{COVID-19} vaccine posts to combat misinformation, help people find where to get the shot \\
\vspace{-2.5mm}
\colorbox{blue!30}{Trump} Gives Delusional \colorbox{green!30}{COVID-19} Press Conference Like A Broken Man \\
\vspace{-2.5mm}
Florida Scientist Says She Was Fired For Not Manipulating \colorbox{green!30}{COVID-19} Data \\
\vspace{-2.5mm}
Wisconsin pharmacist arrested for sabotaging \colorbox{green!30}{COVID-19} vaccines \\
\vspace{-2.5mm}
\colorbox{yellow!30}{Facebook} openly conspiring with FBI to punish \colorbox{blue!30}{Trump} supporters for wrongthink \\
\bottomrule
\end{tabular}
}
\caption{Example of demonstration selection using Algorithm \ref{alg:ideology_estimation} for a given text. Colors show how demonstrations cover salient aspects of the samples.}
\label{tab:bertscore}\vspace{-5mm}
\end{table}

\subsection{Demonstration Selection}
We select $k$ demonstrations for in-context learning (ICL) to augment the LLM's performance.
We utilize \textsc{Set-Cover} \cite{Vazirani2003} to maximize coverage across the candidate pool, where BertScore-Recall (BSR) is chosen as the coverage metric. This strategy is known as \textsc{Set-BSR}, and has been shown to be successful for ICL with multiple classes \citep{gupta2023coverage}.
%
%
%
Additionally, we ensure that an equal number of demonstrations is selected from each ideology class.
A comparison of random and balanced shot selection can be found in Appendix \ref{appendix:balanced-vs-random} to validate this design choice.


%


\begin{figure*}[t]
  \centering
  \includegraphics[width=0.64\textwidth]{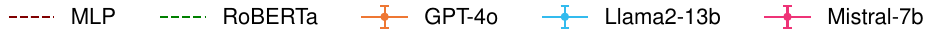}

  \begin{subfigure}[t]{0.32\textwidth}
      \centering
      \includegraphics[width=\textwidth]{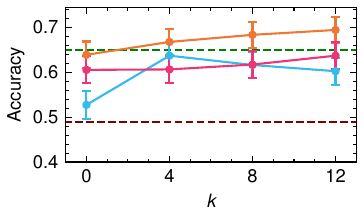}
      \caption{YouTube Slant Dataset}
  \end{subfigure}
  \begin{subfigure}[t]{0.32\textwidth}
      \centering
      \includegraphics[width=\textwidth]{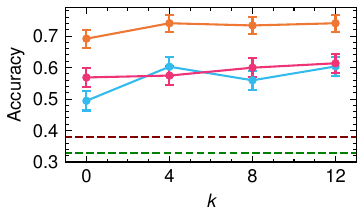}
      \caption{Ad Fontes News Dataset}
  \end{subfigure}
  \begin{subfigure}[t]{0.32\textwidth}
      \centering
      \includegraphics[width=\textwidth]{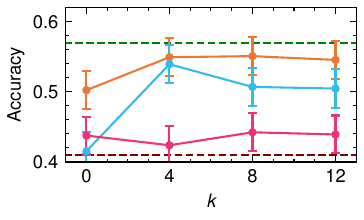}
      \caption{AllSides News Dataset}
  \end{subfigure}
  \caption{Improvement in accuracies of the GPT-4o, Llama2, and Mistral models across the three datasets by increasing the number of ICL demonstrations $k$. The baseline MLP and RoBERTa models are also shown. We see that a higher number of demonstrations leads to better predictions, and that the models generally outperform the baselines (even though the RoBERTa and MLP models have access to the full training set, and the LLMs only see $k$ demonstrations in-context). The error bars show a $95\%$ confidence interval.
  }\vspace{-4mm}
  \label{fig:num_shots}
\end{figure*}

\section{Results and Evaluation}

\subsection{Experimental Setup}
\subsubsection{Datasets}

For our evaluation, we rely on three different datasets, which have been used in prior work on ideology classification.\footnote{While we use a conservatively large candidate set size, we show in Appendix \ref{appendix:candidate-set-size} that even a smaller candidate set size yields similar results.} 

\noindent
\textbf{YouTube Slant Dataset.} 
We sample 645,187 videos from the YouTube dataset by \citet{haroon2023auditing} which estimates the ideology of videos using the Twitter-based slant estimation approach \cite{le2017scalable,barbera2015birds,wojcieszak2023polarized} on a scale of -1 (most liberal) to +1 (most conservative) and was collected between 2021 and 2022.
We use a BERT-based political classifier from \citet{askari2024incentivizing} to further identify 49,894 political videos to create our YouTube slant dataset.
Videos with slant $\leq -0.33$ as were labelled as liberal, $\geq +0.33$ as conservative, and everything else as neutral.
The training set used for selection consists of 48,895 videos from which 20,000 candidates were selected and ranked using the \textsc{Set-BSR} approach.
The test set consists of 999 videos sampled as to maintain an equal balance of ideologies.
Overall, we have videos from over 8,523 channels across the entire training and testing sets (See Appendix \ref{appendix:slant-cutoffs} for where the slant cut-offs lie respective to popular news channels.).

\noindent
\textbf{Ad Fontes News Articles.}
This dataset consists of news articles that have been labeled by expert human annotators on Ad Fontes Media Bias Panel \cite{otero2019ad} from ``most extreme left'' to ``most extreme right.''
Each article is rated by at least three experts from across the political spectrum on a scale of -42 to +42 corresponding to extremely left and extremely right news articles, respectively.
Additionally, Ad Fontes labels consider articles with scores lower than -14.0 to skew left and articles with scores greater than +14.0 as skewing right.
Adopting these cut-offs, we label articles with scores lower than -14.0 as Liberal, scores greater than +14.0 as Conservative, and everything in-between as Neutral.
The training set consists of 33,396 news articles from which we build a candidate set of 8,000 using the \textsc{Set-BSR} approach.
The test set contains 999 news articles with a balanced distribution of ideology.

\noindent
\textbf{AllSides News Articles.}
The dataset contains expert human annotations from AllSides\footnote{\url{http://www.allsides.com/media-bias/media-bias-rating-methods}}, which relies on several human annotation approaches to rank news articles such as blind surveys, expert reviews (on a scale of -6 to +6), and independent reviews.
We use the training and test set of news articles used by \citet{baly2019ideology} from this dataset.
The training set consists of 26,590 news articles, from which we build a candidate set of 8,000 using the \textsc{Set-BSR} approach. The test set consists of 1,300 videos.

\subsubsection{LLMs}
We use three different publicly available LLMs for the experiments.

\noindent \textbf{OpenAI GPT-4o.} We use the pre-trained GPT-4o model by OpenAI \cite{hurst2024gpt} using a paid API subscription. 

\noindent \textbf{Meta Llama-2}. We use the Hugging Face library to run Meta's Llama-2 model with 13B \cite{touvron2023llama} parameters on a local machine with a RTX A6000 GPU, 256GB memory, and 48 cores.

\noindent \textbf{MistralAI Mistral.} We use the Hugging Face library to run MistralAI's Mistral model with 7B parameters \cite{jiang2023mistral} on the same local machine with a RTX A6000 GPU, 256GB memory, and 48 cores.

\subsubsection{Baselines}
We use two standard classification approaches as our supervised learning baselines.

\noindent
\textbf{Multilayer Perceptron (MLP).}
This model is a fully connected neural network that takes a 384-dimension text embedding as input to an initial dense hidden layer of 512 neurons with the \texttt{tanh} activation function, and a three output \texttt{softmax} final layer (for the three ideologies).

\noindent
\textbf{RoBERTa.}
As in \citet{baly2020detect}, we use the base RoBERTa transformer model and fine-tune on each of the three full training datasets for 50 epochs. Unlike the LLM which only receives a small number of $k$ demonstrations, the RoBERTa model sees the full dataset(s).

\begin{figure*}[t!]
  \centering
  \includegraphics[width=0.48\textwidth]{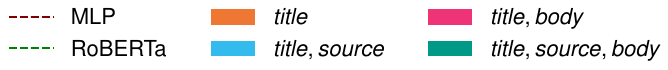}

  \begin{subfigure}[t]{0.32\textwidth}
      \centering
      \includegraphics[width=\textwidth]{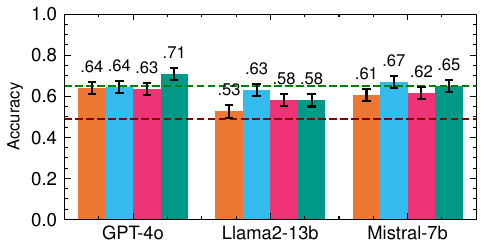}
      \caption{YouTube Slant Dataset}
  \end{subfigure}
  \begin{subfigure}[t]{0.32\textwidth}
      \centering
      \includegraphics[width=\textwidth]{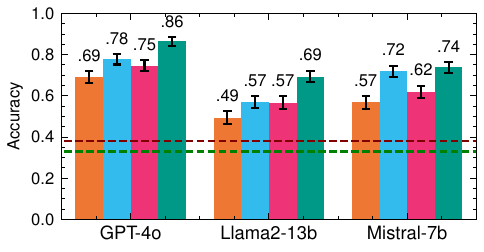}
      \caption{Ad Fontes News Dataset}
  \end{subfigure}
  \begin{subfigure}[t]{0.32\textwidth}
      \centering
      \includegraphics[width=\textwidth]{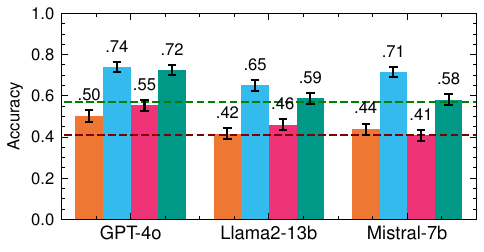}
      \caption{AllSides News Dataset}
  \end{subfigure}
  \caption{
    Improvement in the accuracies of the GPT-4o, Llama2, and Mistral models by testing different combinations of the title, source, and description from the three datasets in the zero-shot setting. We see that providing source and description immensely improves the accuracy of the models across the Ad Fontes and AllSides news datasets where the news sources are more likely to be already known by the model. The YouTube dataset, on the other hand, sees a degradation in performance albeit a slight improvement in the case of Mistral. The error bars show a 95\% confidence interval.
  }\vspace{-4mm}
  \label{fig:accuracy_bar}
\end{figure*}

\subsection{Results}

\noindent
\textbf{Baselines.}
The baseline MLP model performed the worst, achieving an accuracy of 0.49, 0.38, and 0.41 across the YouTube, Ad Fontes, and AllSides datasets.
The RoBERTa baseline achieved a score of 0.65 on the YouTube Dataset and 0.57 on the AllSides News dataset, marked improvements from the MLP model, but fared worse on the Ad Fontes dataset, with accuracy of 0.33.
This clearly shows that simpler models are not well-suited to the task of ideological estimation, especially when compared to more advanced transformer-based architectures.

\noindent
\textbf{Zero-Shot Learning.}
Figure \ref{fig:num_shots} shows the performance of the three LLMs and baselines across the three datasets.
Focusing on the zero-shot setting, we notice that GPT-4o outperformed both Mistral and Llama2, achieving an accuracy of 0.64, 0.69, and 0.50 across the YouTube, Ad Fontes, and AllSides datasets.
Mistral performed better than Llama2, achieving a 0.59 accuracy on the YouTube dataset closer to GPT-4o whereas Llama2's accuracy was very low at 0.44.
On the Ad Fontes dataset, GPT-4o performed 14\% better than Mistral at 0.69 accuracy compared to Mistral's 0.55.
Finally, all LLMs performed poorly on the AllSides dataset with GPT-4o performing the best at 0.50, 7\% lower than the baseline RoBERTa model. 
Comparing with the baselines, RoBERTa performed slightly better than GPT-4o at 0.65 and 0.57 on the YouTube Slant and AllSides datasets, respectively.
In contrast, the MLP performed the worst in the Ad Fontes and AllSides datasets but fared better than Llama2 in the YouTube slant dataset.

\noindent
\textbf{Few-Shot ICL Comparison.} 
Providing the set of demonstrations $\mathcal{D}$ selected using the \textsc{Set-BSR} approach to the prompt $\mathcal{P}$, Figure \ref{fig:num_shots} also shows how the accuracy of the models improved by increasing the number of demonstrations $k$.
The trend of GPT-4o outperforming Llama2 and Mistral persisted as the number of demonstrations increased. GPT-4o consistently achieved the highest scores across all datasets, suggesting that it is particularly well-suited for ideological estimation tasks.

\looseness-1 Mistral-7B and GPT-4o showed a steady increase in performance as the number of demonstrations increased, both seeing a 5\% increase in accuracy going from $k=0 \rightarrow 12$ (0.64 to 0.69 for GPT-4o, and 0.59 to 0.64 for Mistral-7B). Llama2-13B saw marked increases in accuracy across all datasets from $k=0 \rightarrow 4$, but grew worse as $k=4 \rightarrow 12$.

\noindent
\textbf{Providing Additional Metadata.}
Augmenting the test instance $x$ to include source and description increased the performance of the models across all the datasets in the zero-shot setting.
Figure \ref{fig:accuracy_bar} shows how model performance was affected by altering $x$ to include the title, the title and source, the title and description, and the title, source, and description.
For the YouTube dataset, performance for GPT-4o increased by 7\% when the source and description was provided along with the title.
Llama2 saw an increase of 5\% when the description was provided and Mistral saw an increase of 8\% when only the source was provided.
When augmenting the title with the source and description, we see higher increases for Ad Fontes, with the performance of GPT-4o increasing by 17\%, Llama2 by 21\%, and Mistral by 19\%.
Similar increases are seen for AllSides, where adding just the source pushed the accuracy of the models further.
In short, when provided the source information, all three LLMs were able to surpass the baseline RoBERTa model in the zero-shot setting.


\noindent
\textbf{Classifying non-political content.}
Our findings so far have focused on exclusively political content.
The improvement in the accuracy of the LLMs when the source is provided suggests that the source carries substantial weight in determining the ideology of content, where the title alone was previously misclassified.
This improvement is more evident in the Ad Fontes and AllSides news datasets, where the sources are well-known news organizations, and less so in the YouTube slant dataset, where the sources can be any YouTube channel.
To pinpoint the relative role of the title and source in determining the classification, we resample the entire YouTube slant dataset to procure three datasets: 1) political content from well-known news channels (political news), 2) non-political content from well-known news channels (non-political news), and 3) political news from non-news channels, such as influencers or pundits (political non-news).
We use the same training set for the YouTube slant dataset and estimate ideology using GPT-4o.

\begin{figure*}
  \centering
  \begin{subfigure}{0.32\textwidth}
    \includegraphics[width=\textwidth]{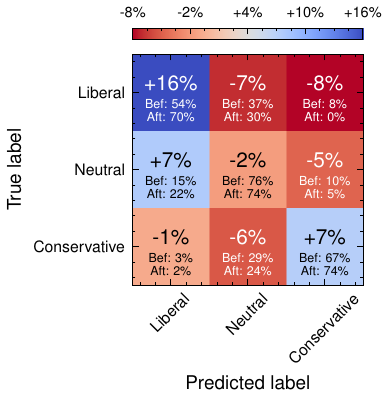}
    \caption{Political news videos}
    \label{fig:pol-news-cm}
  \end{subfigure}
  \begin{subfigure}{0.32\textwidth}
    \includegraphics[width=\textwidth]{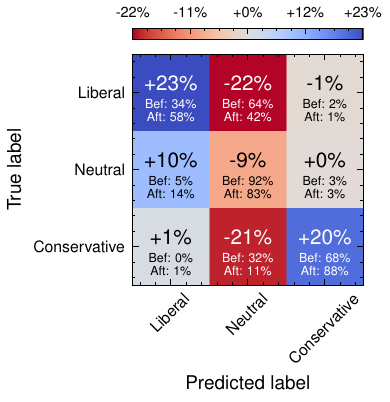}
    \caption{Non-political news videos}
    \label{fig:non-pol-news-cm}
  \end{subfigure}
  \begin{subfigure}{0.32\textwidth}
    \includegraphics[width=\textwidth]{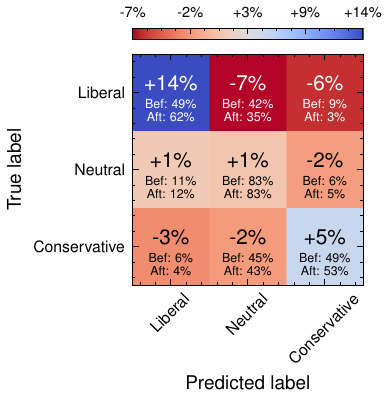}
    \caption{Political non-news videos}
    \label{fig:pol-non-news-cm}
  \end{subfigure}\vspace{-3mm}
  \caption{Heatmap showing changes in the performance of GPT-4o when the channel name is provided for a) political news, b) non-political news, and c) political non-news YouTube videos. Positive values along the main diagonal represent an increase in accuracy. For example, +16\% in the top-left cell means providing the channel name increased the accuracy of classifying the Liberal class from 54\% to 70\%. We see more accurate prediction (increase in top-left and bottom-right) of news content (\ref{fig:pol-news-cm}, \ref{fig:non-pol-news-cm}) when the channel name is provided showing how the LLM is already familiar with the partisan leanings of online news channels.
  }
  \label{fig:pol-chan}\vspace{-6mm}
\end{figure*}

\begin{table}[h]
\centering
\resizebox{\linewidth}{!}{
\begin{tabular}{lccc}
\toprule
& Political & Political & Non-political \\ 
& news & non-news & news \\ 
\midrule
Liberal & 314 & 309 & 125  \\
Neutral & 435 & 436 & 207 \\
Conservative & 251 & 255 & 76 \\
\midrule
Total & 1,000 & 1,000 & 408 \\
\bottomrule
\end{tabular}
}
\caption{Number of samples in the political news, political non-news, and non-political news test sets taken from the YouTube slant dataset. There are fewer examples in the non-political news datasets because that was the maximum number of samples that met this criteria. The training and candidate sets were the same as for the larger YouTube slant dataset.}
\label{tab:pol-news-statistics.tex}
\end{table}
Breaking this down by the class labels, Figure \ref{fig:pol-chan} shows the change in prediction accuracy after adding source information using a heatmap of changes across the three datasets.
The heatmap shows which corresponding class was a given label misclassified as before providing the source allowing us to analyze the influence of the source when estimating ideology.
Coupling this with qualitative examples from Appendix \ref{appendix:qualitative-examples} provides insight into the influence of the source in classifying ideology.

For political news, we see an improvement in accuracy of content previously misclassified as neutral to liberal (+16\%) and conservative (+7\%), becoming almost 70\% more accurate in both cases.
When looking at the kinds of titles re-classified in Table \ref{tab:qual-polnews} in Appendix \ref{appendix:qualitative-examples}, we see that the titles themselves do not indicate the any ideological leaning.
For example, \textit{``Haberman says Trump has been diminished but isn’t dead politically''} was classified as neutral but when the LLM was informed it was from CNN, it was re-classified as liberal.
Similarly, the video titled \textit{``Transgender Clinical Psychologist on people de-transitioning | Dr. Erica Anderson''} was re-classified as conservative when the LLM learned it came from Newsmax, a conservative outlet, despite the title itself indicating no such political leaning.
Cases where the label changed from liberal to conservative or vice-versa showed more nuance. 
The video titled \textit{``Democrats introduce bill to block Trump from holding office''} was classified as liberal but then re-classified as conservative when the source Newsmax was provided which was consistent with the content of the video.
This is true for the non-political news dataset as well, where the titles do not depict any political inclination as can be seen in Table \ref{tab:qual-nonpolnews} in Appendix \ref{appendix:qualitative-examples} but the model re-classified it according to the ideology of the news source.
Finally, for the political non-news videos, we do not see much movement as the LLM is likely unfamiliar with the sources provided.
The cases where we do see movement as seen in Table \ref{tab:qual-polnonnews}, it is also driven by the source of the video as well.
We see some movement towards the neutral as well.
For example, the videos titled \textit{``The Trouble with the Electoral College''} and \textit{``Is capitalism actually broken?''} are liberal in ideology but when the model learns they are from educative channels CGP Grey and TED-Ed, it re-classifies them as neutral.

\subsection{Discussion}


\noindent
\textbf{Few-shot ICL increases performance.}
With higher number of shots, we see that  accuracy improves across the board for all three LLMs across all three datasets.
Except for Llama2-13B which performs best at 4-shots (likely due to its shorter context window length), both GPT-4o and Mistral-7B see continued improvement in accuracy at an average of 5\% and 3\% across all three datasets.

\noindent
\textbf{Supplementing prompt with additional metadata improves performance.}
The additional metadata consist of the source and description of the content.
Just adding the content source increased the accuracy of all the models, suggesting that the models are familiar with the ideological positions of the sources used and leverage that information to improve their ideology estimation.
In contrast, the description actually lowers the accuracy when added, likely because it affects the context-window of the input poorly and may not provide more context than the title itself.

\noindent
\textbf{Leveraging few-shot ICL and additional metadata for best performance.}
Combining the few-shot ICL $k=12$ and incorporating the source resulted in the highest gains in GPT-4o's performance across all three datasets.
The model achieved a 10\% increase in the YouTube slant dataset ($0.64 \rightarrow 0.74$), 18\% increase in the Ad Fontes dataset ($0.69 \rightarrow 0.87$), and 37\% increase in the AllSides dataset ($0.50 \rightarrow 0.87$).
These improvements demonstrate how GPT-4o leverages details from relevant demonstrations and its existing knowledge of news sources to more accurately reason about the ideology of content.

\noindent
\textbf{Multi-modal classification using YouTube thumbnails does not improve GPT-4o's performance.}
We experiment with adding video thumbnails as part of the prompt to GPT-4o.
Despite the potential of multi-modal classification methods to enhance model performance by incorporating visual information, our experiments reveal that using YouTube thumbnails as an additional modality does not improve GPT-4o's performance. When comparing the results of GPT-4o with and without the inclusion of YouTube thumbnails, the accuracy of the model dropped to 0.35 from 0.64 on the YouTube slant dataset.
This result indicates that, for the problem of ideology estimation from YouTube content, visual information does not contribute additional meaningful signals beyond what is captured in the text and instead can be misleading for the model. However, more extensive analysis and experiments are needed for incorporating multi-modal information for ideology classification and can be undertaken in future work.
%

\noindent
\textbf{Chain-of-thought reasoning is ineffective.}
We apply chain-of-thought (CoT) reasoning (i.e., prompting the model to think through the task step-by-step) to GPT-4o for our task and see that it doesn't improve accuracy despite its known benefits \cite{kojima2024zeroshot}. 
On the YouTube slant dataset, the model accuracy dropped from 0.64 to 0.62 when CoT reasoning was used.
One explanation for this finding is that ideology estimation is primarily reliant on identifying patterns in content that are explicit rather than requiring deep logical steps. The task might not demand the same kind of multi-step reasoning that CoT typically helps with in applications, such as arithmetic or procedural problem-solving. 

\noindent
\textbf{Accuracy of LLMs is comparable with human annotation.}
The high accuracy of GPT-4o is comparable to human annotation on the AllSides and Ad Fontes News datasets, both of which are human-annotated. 
On these, GPT-4o achieved high accuracy, closely aligning with human judgments, exceeding 0.80 in some configurations.
For additional validation, a domain expert annotated a sample of 561 videos in the YouTube dataset, finding that GPT-4o agreed with the human annotation 79\% of the time. This suggests that GPT-4o is able to effectively mirror human-level ideological classification, even in more informal content. 
The model's performance highlights its capacity to generalize patterns and signals used by human annotators, making it a valuable tool for automated ideological estimation.





\section{Conclusion}
This study explored the efficacy of few-shot learning using Large Language Models for the task of ideology estimation across diverse textual datasets, including news articles and YouTube videos. Our findings demonstrate that LLMs, particularly GPT-4o, achieve state-of-the-art performance, surpassing traditional supervised baselines. By leveraging advanced few-shot in-context learning methods, we addressed challenges of scalability and robustness, showcasing the ability of LLMs to adapt to evolving contexts with minimal supervision.

\section{Limitations}

Despite the promising results and contributions outlined in this paper, several limitations must be acknowledged to identify areas for future work. The performance of our approach is heavily reliant on the capabilities of pre-trained LLMs such as GPT-4o, Llama2-13B, and Mistral-7B. Variability in model updates or differences in underlying pre-training data could affect reproducibility and consistency across different deployments. In addition, while incorporating metadata, such as source information improved performance, it also introduced the potential for biased predictions. Models may disproportionately rely on known ideological leanings of sources, potentially misclassifying nuanced or neutral content. This issue might be further exacerbated by the pre-processing guardrails that serve to filter data, but inadvertently possess their own robustness and bias concerns \citep{achara2025watching, yang2024benchmarking}. 

Furthermore, our experiments focus on text, and extending these methods to classify ideology in non-textual data such as video or audio remains an open challenge and an important direction for future work. 
Lastly, we acknowledge that the datasets used for evaluation, while diverse, may not fully capture the breadth of content and ideological nuance present in real-world social media platforms. For instance, the balance of liberal, neutral, and conservative content in the datasets may not reflect natural distributions, and also the majority of content consumed by users is generated by other users themselves, not by news media organizations or YouTube channels, suggesting the need to adapt our approach to user-generated posts and comments. Finally, while beyond the scope of our work, LLMs can themselves have several robustness issues \citep{askari2025assessing, chhabra2024revisiting} that can result in potential downstream evaluation issues.

\section{Ethics Statement}
Automated ideology estimation carries inherent risks, including misuse for surveillance or censorship. Furthermore, reliance on LLMs can perpetuate biases present in training data, which might influence predictions in unintended ways. While the study evaluates multiple datasets, the generalizability of the approach across all social media platforms, with varying content types and user behaviors, is yet to be fully established.

Addressing these limitations could involve refining LLM prompting techniques, improving model transparency, exploring alternative datasets, and incorporating mechanisms for bias mitigation and real-time adaptability. Future research should also emphasize ethical frameworks for deploying such systems responsibly, ensuring that advancements in technology contribute positively to societal discourse without exacerbating existing challenges.

\bibliography{references}

\clearpage

\appendix
\section*{Appendix}


\section{Prompts used for the LLM}
\label{appendix:task_prompts}

Here, we present the prompts that are used for querying the models.

\vspace{1em}
\noindent \textbf{Prompt for querying with just the title.}
\vspace{0.5em}
\noindent
\fbox{\parbox{0.46\textwidth}{\textit{Classify the following news article titles as ideologically liberal, neutral, or conservative. Titles with no ideological content are classified as neutral. Only respond with the final answer.}}}

\vspace{1em}
\noindent \textbf{Prompt for querying with the title and source.}
\vspace{0.5em}
\noindent
\fbox{\parbox{0.46\textwidth}{\textit{Classify the following news article titles as ideologically liberal, neutral, or conservative. Titles with no ideological content are classified as neutral. \textbf{The news source is also specified for additional context.} Only respond with the final answer.}}}

\vspace{1em}
\noindent \textbf{Prompt for querying with the title, description, and source.}
\vspace{0.5em}
\noindent
\fbox{\parbox{0.46\textwidth}{\textit{Classify the following news article titles as ideologically liberal, neutral, or conservative. Titles with no ideological content are classified as neutral. \textbf{The news description is also specified for additional context.} Only respond with the final answer.}}}

\section{Statistical significances}
\label{appendix:stat-sig}

We use the McNemar test to determine the statistical differences between the accuracies of two specific classifiers.
Tables \ref{tab:sig-youtube-combinations}, \ref{tab:sig-adfontes-combinations}, and \ref{tab:sig-allsides-combinations} show the significance values for the models trained on the YouTube Slant, Ad Fontes news, and AllSides news datasets respectively.

\begin{table*}[h]
\centering
\resizebox{\linewidth}{!}{
\begin{tabular}{lccccccccccccc}
& & \multicolumn{4}{c}{GPT} &  \multicolumn{4}{c}{Llama}  &  \multicolumn{4}{c}{Mistral} \\ 
\cmidrule(lr){3-6}\cmidrule(lr){7-10}\cmidrule(lr){11-14}
& & T & T/S & T/B & T/S/B &T & T/S & T/B & T/S/B &T & T/S & T/B & T/S/B \\
\midrule
\multirow{4}{*}{GPT} & T & --- & 0.13 & 0.06 & 33.06** & 34.18** & 0.21 & 11.05** & 10.75** & 4.50* & 4.21* & 2.34 & 0.39 \\
& T/S & 0.13 & --- & 0.52 & 27.95** & 35.79** & 0.69 & 12.10** & 12.30** & 4.72* & 2.79 & 3.31 & 0.04 \\
& T/B & 0.06 & 0.52 & --- & 42.29** & 30.30** & 0.06 & 10.25** & 10.16** & 2.96 & 5.33* & 1.70 & 0.88 \\
& T/S/B & 33.06** & 27.95** & 42.29** & --- & 89.28** & 22.45** & 59.74** & 58.72** & 41.44** & 7.87** & 45.75** & 22.90** \\
\midrule
\multirow{4}{*}{Llama} & T & 34.18** & 35.79** & 30.30** & 89.28** & --- & 37.09** & 9.03** & 8.62** & 19.19** & 58.08** & 23.16** & 43.91** \\
& T/S & 0.21 & 0.69 & 0.06 & 22.45** & 37.09** & --- & 7.81** & 8.57** & 1.75 & 5.46* & 0.56 & 1.20 \\
& T/B & 11.05** & 12.10** & 10.25** & 59.74** & 9.03** & 7.81** & --- & 0.00 & 1.71 & 26.06** & 4.14* & 16.75** \\
& T/S/B & 10.75** & 12.30** & 10.16** & 58.72** & 8.62** & 8.57** & 0.00 & --- & 1.83 & 26.38** & 4.38* & 17.32** \\
\midrule
\multirow{4}{*}{Mistral} & T & 4.50* & 4.72* & 2.96 & 41.44** & 19.19** & 1.75 & 1.71 & 1.83 & --- & 18.21** & 0.43 & 7.90** \\
& T/S & 4.21* & 2.79 & 5.33* & 7.87** & 58.08** & 5.46* & 26.06** & 26.38** & 18.21** & --- & 13.56** & 2.79 \\
& T/B & 2.34 & 3.31 & 1.70 & 45.75** & 23.16** & 0.56 & 4.14* & 4.38* & 0.43 & 13.56** & --- & 9.75** \\
& T/S/B & 0.39 & 0.04 & 0.88 & 22.90** & 43.91** & 1.20 & 16.75** & 17.32** & 7.90** & 2.79 & 9.75** & --- \\
\bottomrule
\end{tabular}
}
\caption{Statistical significance values using the McNemar test for the YouTube slant dataset. T is Title, S is Source, and B is Body of the post. ** if $p < 0.01$ and * if $p < 0.05$.}
\label{tab:sig-youtube-combinations}
\end{table*}
\begin{table*}[h]
\centering
\resizebox{\linewidth}{!}{
\begin{tabular}{lccccccccccccc}
& & \multicolumn{4}{c}{GPT} &  \multicolumn{4}{c}{Llama}  &  \multicolumn{4}{c}{Mistral} \\ 
\cmidrule(lr){3-6}\cmidrule(lr){7-10}\cmidrule(lr){11-14}
& & T & T/S & T/B & T/S/B &T & T/S & T/B & T/S/B &T & T/S & T/B & T/S/B \\
\midrule
\multirow{4}{*}{GPT}  & T & --- & 34.40** & 20.11** & 449.51** & 96.77** & 32.68** & 38.99** & 224.78** & 51.50** & 2.29 & 18.13** & 288.68** \\
 & T/S & 34.40** & --- & 6.67** & 435.86** & 200.31** & 132.25** & 140.91** & 218.99** & 136.48** & 19.91** & 82.16** & 265.40** \\
 & T/B & 20.11** & 6.67** & --- & 440.16** & 163.22** & 81.73** & 103.88** & 233.55** & 104.43** & 3.12 & 62.26** & 283.78** \\
 & T/S/B & 259.15** & 373.77** & 318.02** & --- & 68.03** & 125.03** & 124.95** & 107.43** & 130.00** & 286.31** & 175.39** & 83.12** \\
\midrule
\multirow{4}{*}{Llama}  & T & 96.77** & 200.31** & 163.22** & 433.77** & --- & 21.99** & 21.16** & 236.90** & 18.25** & 138.91** & 40.15** & 276.51** \\
 & T/S & 32.68** & 132.25** & 81.73** & 475.86** & 21.99** & --- & 0.00 & 274.00** & 0.00 & 85.23** & 5.80* & 299.13** \\
 & T/B & 38.99** & 140.91** & 103.88** & 468.49** & 21.16** & 0.00 & --- & 271.24** & 0.00 & 81.82** & 8.23** & 299.39** \\
 & T/S/B & 252.24** & 383.38** & 327.86** & 107.43** & 79.93** & 148.70** & 144.11** & --- & 136.57** & 313.88** & 184.20** & 7.11** \\
\midrule
\multirow{4}{*}{Mistral}  & T & 51.50** & 136.48** & 104.43** & 451.99** & 18.25** & 0.00 & 0.00 & 237.93** & --- & 86.05** & 9.88** & 282.52** \\
 & T/S & 2.29 & 19.91** & 3.12 & 465.55** & 138.91** & 85.23** & 81.82** & 249.35** & 86.05** & --- & 33.00** & 295.92** \\
 & T/B & 18.13** & 82.16** & 62.26** & 453.61** & 40.15** & 5.80* & 8.23** & 240.67** & 9.88** & 33.00** & --- & 295.82** \\
 & T/S/B & 240.52** & 343.16** & 295.33** & 83.12** & 56.78** & 104.94** & 107.75** & 7.11** & 113.84** & 263.56** & 161.16** & --- \\
\bottomrule
\end{tabular}
}
\caption{Statistical significance values using the McNemar test for the AdFontes news dataset. T is Title, S is Source, and B is Body of the post. ** if $p < 0.01$ and * if $p < 0.05$.}
\label{tab:sig-adfontes-combinations}
\end{table*}
\begin{table*}[h]
\centering
\resizebox{\linewidth}{!}{
\begin{tabular}{lccccccccccccc}
& & \multicolumn{4}{c}{GPT} &  \multicolumn{4}{c}{Llama}  &  \multicolumn{4}{c}{Mistral} \\ 
\cmidrule(lr){3-6}\cmidrule(lr){7-10}\cmidrule(lr){11-14}
& & T & T/S & T/B & T/S/B &T & T/S & T/B & T/S/B &T & T/S & T/B & T/S/B \\
\midrule
\multirow{4}{*}{GPT} & T & --- & 277.85** & 32.27** & 255.80** & 23.89** & 60.47** & 6.33* & 25.05** & 17.31** & 186.27** & 47.23** & 33.34** \\
 & T/S & 277.85** & --- & 195.62** & 3.90* & 258.18** & 26.08** & 212.25** & 81.25** & 272.09** & 4.23* & 357.08** & 139.18** \\
 & T/B & 32.27** & 195.62** & --- & 192.83** & 56.81** & 26.50** & 30.13** & 3.88* & 52.94** & 118.53** & 107.43** & 4.20* \\
 & T/S/B & 255.80** & 3.90* & 192.83** & --- & 240.60** & 17.80** & 200.62** & 70.24** & 253.02** & 0.56 & 342.79** & 124.91** \\
\midrule
\multirow{4}{*}{Llama} & T & 23.89** & 258.18** & 56.81** & 240.60** & --- & 247.06** & 12.31** & 132.97** & 2.17 & 260.91** & 0.13 & 80.77** \\
 & T/S & 60.47** & 26.08** & 26.50** & 17.80** & 247.06** & --- & 164.01** & 24.63** & 162.56** & 17.64** & 151.72** & 16.56** \\
 & T/B & 6.33* & 212.25** & 30.13** & 200.62** & 12.31** & 164.01** & --- & 140.34** & 2.27 & 204.32** & 10.70** & 50.39** \\
 & T/S/B & 25.05** & 81.25** & 3.88* & 70.24** & 132.97** & 24.63** & 140.34** & --- & 89.82** & 71.34** & 108.30** & 0.17 \\
\midrule
\multirow{4}{*}{Mistral} & T & 17.31** & 272.09** & 52.94** & 253.02** & 2.17 & 162.56** & 2.27 & 89.82** & --- & 292.91** & 4.31* & 87.31** \\
 & T/S & 186.27** & 4.23* & 118.53** & 0.56 & 260.91** & 17.64** & 204.32** & 71.34** & 292.91** & --- & 339.67** & 138.56** \\
 & T/B & 47.23** & 357.08** & 107.43** & 342.79** & 0.13 & 151.72** & 10.70** & 108.30** & 4.31* & 339.67** & --- & 175.10** \\
 & T/S/B & 33.34** & 139.18** & 4.20* & 124.91** & 80.77** & 16.56** & 50.39** & 0.17 & 87.31** & 138.56** & 175.10** & --- \\
\bottomrule
\end{tabular}
}
\caption{Statistical significance values using the McNemar test for the AllSides news dataset. T is Title, S is Source, and B is Body of the post. ** if $p < 0.01$ and * if $p < 0.05$.}
\label{tab:sig-allsides-combinations}
\end{table*}

\section{Sources can be misleading}
\label{appendix:sources-misleading}

Our prior results showed that providing the source information substantially increases the accuracy.
However, not all content from a given source may share the ideology of the source itself \cite{haroon2023auditing}, as some liberal articles can come from conservative outlets.
To highlight this particular scenario, we identify cases in our test set where the ideology of the article or video is different from the ideology of its source for all three datasets (e.g., a liberal article from Fox News or a conservative article from MSNBC).
We rely on AllSides' media bias chart\footnote{https://www.allsides.com/media-bias/media-bias-chart} for source ideology and map these sources to their corresponding news articles and YouTube videos.
We then compare the accuracy of GPT-4o when 1) only title is provided, 2) only source is provided, and 3) both title and source are provided.

\begin{table}[h]
\centering
\resizebox{\linewidth}{!}{
\begin{tabular}{ll}
\toprule
\textbf{Liberal Sources} & \textbf{Conservative Sources} \\
\midrule
ABC News & Breitbart\\
Associated Press& CBN\\
CBC News& Daily Caller\\
CBS News& Daily Mail\\
CNN& Daily Wire\\
Daily Beast& Fox Business\\
HuffPost& Fox News\\
Jacobin& New York Post\\
MSNBC& Newsmax\\
Mother Jones& The American Conservative\\
NPR& The American Spectator\\
Slate& The Blaze\\
The Atlantic& The Daily Caller\\
The Guardian& The Epoch Times\\
The Intercept& The Federalist\\
The Nation& The Federalist Society\\
The New York Times& The Post Millennial\\
The New Yorker& The Washington Free Beacon\\
Vox& Washington Examiner\\
\bottomrule
\end{tabular}
}
\caption{Liberal and conservative sources considered for the misleading sources experiment.}
\label{tab:liberal-conservative-sources}
\end{table}

Table \ref{tab:liberal-conservative-sources} lists the names of the left and right sources identified and mapped to the three datasets.
In the YouTube test set, we identified 40 cases where an neutral/conservative article from a liberal source and 21 cases where a liberal/neutral article came from a conservative source.
The Ad Fontes news test set had 38 cases of neutral/conservative articles from liberal sources and 11 cases of liberal/neutral articles from conservative sources.
Finally, the AllSides news test set had 98 neutral/conservative articles from liberal sources and 0 liberal/neutral articles from conservative sources.

\begin{table}[h]
\centering
\resizebox{\linewidth}{!}{

\begin{tabular}{ccccc}
\toprule
\multicolumn{5}{c}{Non-liberal articles from liberal sources} \\
\toprule
Title & Source & YouTube Slant & Ad Fontes News & AllSides News \\
\midrule
\cmark &  & 0.63 & 0.58 & 0.49 \\
 & \cmark & 0.47 & 0.38 & 1.00 \\
\cmark & \cmark & 0.66 & 0.70 & 1.00 \\
\toprule
\toprule
\multicolumn{5}{c}{Non-conservative articles from conservative sources} \\
\toprule
Title & Source & YouTube Slant & Ad Fontes News & AllSides News \\
\midrule
\cmark &  & 0.45 & 0.57 & --- \\
 & \cmark & 0.00 & 0.00 & --- \\
\cmark & \cmark & 0.00 & 0.52 & --- \\
\bottomrule
\end{tabular}
}

\caption{Accuracy of GPT-4o on test set where only non-liberal/non-conservative articles were considered from liberal/conservative sources.}
\label{tab:liberal-conservative-accuracies}
\end{table}

Table \ref{tab:liberal-conservative-accuracies} reports the accuracy of the LLMs on these specific test cases for liberal and conservative sources respectively.
We see that the LLMs suffer from incorrect classifications when only the source is considered as the ground truth contradicts the ideology of the source.
Here, providing the title increases the accuracy of the model especially in the conservative sources where the accuracy on just the source is $0\%$ for the YouTube and Ad Fontes datasets.
In fact, providing the title for the YouTube test set in this case still does not improve the accuracy seeing as the model accuracy is still $0\%$. This suggests that the LLM overtly relies on the source in this case completely disregarding title wherein accuracy on just the title was close to $45\%$.
Similar trends can be seen for the Ad Fontes news test set where the accuracy of source alone was $38\%$ and title was $58\%$, and both was $70\%$.
We observe that the AllSides news test set still has better performance when the source is provided at a $100\%$.
On closer inspection, we observe that in the test set sampled, there is a high correlation between the source and article ideology which likely leads to such a high performance when source is provided.

\section{Few Shot Results for all Datasets}
\begin{figure*}[t!]
  \centering
  \begin{subfigure}[t]{0.32\textwidth}
      \centering
      \includegraphics[width=\textwidth]{figures/num_shots_ytideology.pdf}
      \caption{YouTube Slant Dataset}
  \end{subfigure}
  \hfill
  \begin{subfigure}[t]{0.32\textwidth}
      \centering
      \includegraphics[width=\textwidth]{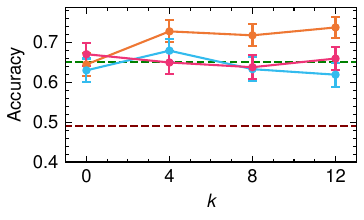}
      \caption{YouTube Slant Dataset w/ Channel}
  \end{subfigure}
  \hfill
  \begin{subfigure}[t]{0.32\textwidth}
      \centering
      \includegraphics[width=\textwidth]{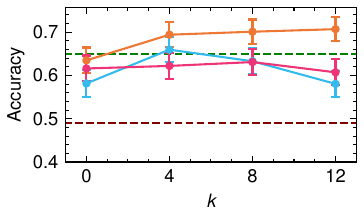}
      \caption{YouTube Slant Dataset w/ Description}
  \end{subfigure}

  \begin{subfigure}[t]{0.32\textwidth}
    \centering
    \includegraphics[width=\textwidth]{figures/num_shots_newsideology.pdf}
    \caption{News Ideology Dataset}
\end{subfigure}
\hfill
\begin{subfigure}[t]{0.32\textwidth}
    \centering
    \includegraphics[width=\textwidth]{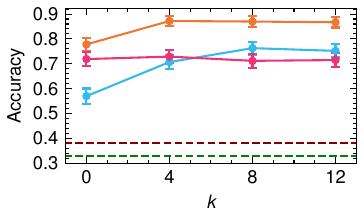}
    \caption{News Ideology Dataset w/ Source}
\end{subfigure}
\hfill
\begin{subfigure}[t]{0.32\textwidth}
    \centering
    \includegraphics[width=\textwidth]{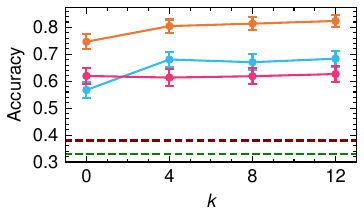}
    \caption{News Ideology Dataset w/ Description}
\end{subfigure}

\begin{subfigure}[t]{0.32\textwidth}
  \centering
  \includegraphics[width=\textwidth]{figures/num_shots_balynewsideology.pdf}
  \caption{Baly Dataset}
\end{subfigure}
\hfill
\begin{subfigure}[t]{0.32\textwidth}
  \centering
  \includegraphics[width=\textwidth]{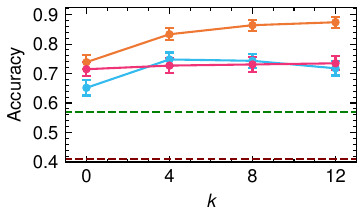}
  \caption{Baly Dataset w/ Source}
\end{subfigure}
\hfill
\begin{subfigure}[t]{0.32\textwidth}
  \centering
  \includegraphics[width=\textwidth]{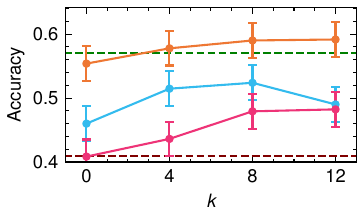}
  \caption{Baly Dataset w/ Description}
\end{subfigure}
\caption{Increases in accuracy across all datasets by increasing the number of ICL shots.}
\label{fig:increases-all-datasets}
\end{figure*}

Figure \ref{fig:increases-all-datasets} shows the increases in accuracy across all datasets by increasing the number of shots.

\section{Ablations on different parameters}
\label{appendix:ablations}

In this section, we report on the various ablations of title, source, description, and number of shots for the various baseline and LLM models.

\subsection{MLP Baseline}

Table \ref{tab:ablations-mlp} shows the accuracy of the MLP model under different title, source, and description ablations.
The multi-layer perceptron model achieved good accuracy when the source was provided.
In the AllSides news dataset, wherein we observe that the source is highly correlated with article ideology, the model achieves $69\%$ accuracy when just the source is provided.
Similarly, MLP achieves $64\%$ accuracy on the YouTube slant dataset when title, source, and article description are provided.
Finally, the model achieves a high accuracy of $62\%$ on the Ad Fontes news dataset when just the title and source are provided.

\begin{table}[h!]
\centering
\resizebox{\linewidth}{!}{
\begin{tabular}{cccccc}
\toprule
 & &  & YouTube & Ad Fontes & AllSides \\
 Title & Source & Body & Slant & News & News \\
\midrule
\cmark &  &  & 0.51 & 0.38 & 0.43 \\
& \cmark &  & 0.61 & 0.55 & 0.69 \\
&  & \cmark & 0.55 & 0.38 & 0.47 \\
\cmark & \cmark &  & 0.63 & 0.62 & 0.57 \\
\cmark &  & \cmark & 0.57 & 0.43 & 0.46 \\
& \cmark & \cmark & 0.62 & 0.60 & 0.58 \\
\cmark & \cmark & \cmark & 0.64 & 0.61 & 0.52 \\
\bottomrule
\end{tabular}
}
\caption{Ablations on the baseline multi-layer perceptron model on the three datasets.}
\label{tab:ablations-mlp}
\end{table}

\subsection{RoBERTa baseline}

Table \ref{tab:ablations-roberta} shows the accuracy of the RoBERTa model under different title, source, and description ablations.
The RoBERTa model performed well on the YouTube slant dataset achieving an accuracy of $65\%$ when title, source, and description were all provided.
The performance was worse on the news datasets, with a maximum of $48\%$ on the Ad Fontes news dataset and $58\%$ on the AllSides news dataset.

\begin{table}[h!]
\centering
\resizebox{\linewidth}{!}{
\begin{tabular}{cccccc}
\toprule
  & &  & YouTube & Ad Fontes & AllSides \\
  Title & Source & Body & Slant & News & News \\
\midrule
\cmark &  &  & 0.57 & 0.35 & 0.52 \\
& \cmark &  & 0.45 & 0.48 & 0.48 \\
&  & \cmark & 0.62 & 0.35 & 0.50 \\
\cmark & \cmark &  & 0.65 & 0.44 & 0.58 \\
\cmark &  & \cmark & 0.64 & 0.38 & 0.53 \\
& \cmark & \cmark & 0.65 & 0.44 & 0.58 \\
\cmark & \cmark & \cmark & 0.65 & 0.44 & 0.55 \\
\bottomrule
\end{tabular}
}
\caption{Ablations on the baseline RoBERTa model on the three datasets.}
\label{tab:ablations-roberta}
\end{table}

\subsection{GPT-4o}

Table \ref{tab:ablations-gpt4o} shows the accuracy of the GPT-4o model under different ablations of title, source, and description across the three datasets and varying number of demonstrations $k$ considered.
We see that the model performs the best when given a higher number of demonstrations for the YouTube slant and Ad Fontes news datasets.
For the YouTube dataset, the model performed best when title, source, and description were all provided whereas for the Ad Fontes news dataset, only title and source were sufficient.
The AllSides news dataset showed a different trend in that zero demonstrations gave the best score when only the source was provided.
We observed in the test set for this dataset that source and target label were highly correlated (i.e., a video from Fox,  a well-known conservative outlet, was almost always labeled conservative) making it simpler for the model to just assign the source ideology as the target ideology.
In Appendix \ref{appendix:sources-misleading}, we investigate this further by considering edge cases in the test set.

\begin{table}[h!]
\centering
\resizebox{\linewidth}{!}{
\begin{tabular}{ccccccc}
\toprule
& & &  & YouTube & Ad Fontes & AllSides \\
$k$ & Title & Source & Body & Slant & News & News \\
\midrule
0 & \cmark &  &  & 0.64 & 0.69 & 0.50 \\
0 &  & \cmark &  & 0.63 & 0.80 & 0.88 \\
0 &  &  & \cmark & 0.56 & 0.64 & 0.44 \\
0 & \cmark & \cmark &  & 0.64 & 0.78 & 0.74 \\
0 & \cmark &  & \cmark & 0.63 & 0.75 & 0.55 \\
0 &  & \cmark & \cmark & 0.59 & 0.82 & 0.85 \\
0 & \cmark & \cmark & \cmark & 0.71 & 0.86 & 0.72 \\
\midrule
4 & \cmark &  &  & 0.67 & 0.74 & 0.55 \\
4 & \cmark & \cmark &  & 0.73 & 0.87 & 0.83 \\
4 & \cmark &  & \cmark & 0.69 & 0.80 & 0.58 \\
4 & \cmark & \cmark & \cmark & 0.72 & 0.86 & 0.82 \\
\midrule
8 & \cmark &  &  & 0.68 & 0.73 & 0.55 \\
8 & \cmark & \cmark &  & 0.72 & 0.87 & 0.86 \\
8 & \cmark &  & \cmark & 0.70 & 0.81 & 0.59 \\
8 & \cmark & \cmark & \cmark & 0.71 & 0.86 & 0.84 \\
\midrule
12 & \cmark &  &  & 0.69 & 0.74 & 0.55 \\
12 & \cmark & \cmark &  & 0.74 & 0.87 & 0.87 \\
12 & \cmark &  & \cmark & 0.71 & 0.82 & 0.59 \\
12 & \cmark & \cmark & \cmark & 0.71 & 0.87 & 0.84 \\
\bottomrule
\end{tabular}
}
\caption{Ablations on the GPT-4o on the three datasets. Here, $k$ is the number of demonstrations.}
\label{tab:ablations-gpt4o}
\end{table}

\subsection{Llama2}

Table \ref{tab:ablations-llama} shows the accuracy of the Llama2 model under different ablations of title, source, and description across the three datasets and varying number of demonstrations $k$ considered.
We see that the model performs the best at 4 or fewer demonstrations indicating that more demonstrations leads to a drop in performance likely due to a limited context window.
For the YouTube and Ad Fontes news datasets, the model performed best when title, source, and only four demonstrations were provided.
The AllSides news dataset showed similar trends as GPT-4o in that the source was sufficient without any demonstrations to achieve the best accuracy.

\begin{table}[h!]
  \centering
  \resizebox{\linewidth}{!}{
  \begin{tabular}{ccccccc}
  \toprule
  & & &  & YouTube & Ad Fontes & AllSides \\
  $k$ & Title & Source & Body & Slant & News & News \\
  \midrule
  0 & \cmark &  &  & 0.53 & 0.49 & 0.42 \\
0 &  & \cmark &  & 0.61 & 0.52 & 0.82 \\
0 &  &  & \cmark & 0.52 & 0.49 & 0.37 \\
0 & \cmark & \cmark &  & 0.63 & 0.57 & 0.65 \\
0 & \cmark &  & \cmark & 0.58 & 0.57 & 0.46 \\
0 &  & \cmark & \cmark & 0.48 & 0.65 & 0.71 \\
0 & \cmark & \cmark & \cmark & 0.58 & 0.69 & 0.59 \\
\midrule
4 & \cmark &  &  & 0.64 & 0.60 & 0.54 \\
4 & \cmark & \cmark &  & 0.68 & 0.79 & 0.75 \\
4 & \cmark &  & \cmark & 0.66 & 0.68 & 0.51 \\
4 & \cmark & \cmark & \cmark & 0.64 & 0.79 & 0.68 \\
\midrule
8 & \cmark &  &  & 0.60 & 0.56 & 0.51 \\
8 & \cmark & \cmark &  & 0.63 & 0.76 & 0.74 \\
8 & \cmark &  & \cmark & 0.63 & 0.67 & 0.52 \\
8 & \cmark & \cmark & \cmark & 0.60 & 0.78 & 0.67 \\
\midrule
12 & \cmark &  &  & 0.60 & 0.60 & 0.50 \\
12 & \cmark & \cmark &  & 0.62 & 0.75 & 0.72 \\
12 & \cmark &  & \cmark & 0.58 & 0.68 & 0.49 \\
12 & \cmark & \cmark & \cmark & 0.56 & 0.78 & 0.67 \\
  \bottomrule
  \end{tabular}
  }
  \caption{Ablations on the Llama2 on the three datasets. Here, $k$ is the number of demonstrations.}
  \label{tab:ablations-llama}
  \end{table}
  
\subsection{Mistral}

Table \ref{tab:ablations-mistral} shows the accuracy of the Mistral model under different ablations of title, source, and description across the three datasets and varying number of demonstrations $k$ considered.
We see that the model performs the best at zero demonstrations indicating that more demonstrations leads to a drop in performance likely due to a limited context window.
For the YouTube and Ad Fontes news datasets, the model performed best when title, source, and description were provided with zero demonstrations.
The AllSides news dataset again showed similar trends as GPT-4o in that the source was sufficient without any demonstrations to achieve the best accuracy.

\begin{table}[h!]
  \centering
  \resizebox{\linewidth}{!}{
  \begin{tabular}{ccccccc}
  \toprule
  & & &  & YouTube & Ad Fontes & AllSides \\
  $k$ & Title & Source & Body & Slant & News & News \\
  \midrule
  0 & \cmark &  &  & 0.61 & 0.57 & 0.44 \\
0 &  & \cmark &  & 0.61 & 0.72 & 0.83 \\
0 &  &  & \cmark & 0.52 & 0.59 & 0.34 \\
0 & \cmark & \cmark &  & 0.67 & 0.72 & 0.71 \\
0 & \cmark &  & \cmark & 0.62 & 0.62 & 0.41 \\
0 &  & \cmark & \cmark & 0.57 & 0.73 & 0.66 \\
0 & \cmark & \cmark & \cmark & 0.65 & 0.74 & 0.58 \\
\midrule
4 & \cmark &  &  & 0.61 & 0.57 & 0.42 \\
4 & \cmark & \cmark &  & 0.65 & 0.73 & 0.73 \\
4 & \cmark &  & \cmark & 0.62 & 0.61 & 0.44 \\
4 & \cmark & \cmark & \cmark & 0.63 & 0.68 & 0.60 \\
\midrule
8 & \cmark &  &  & 0.62 & 0.59 & 0.44 \\
8 & \cmark & \cmark &  & 0.64 & 0.71 & 0.73 \\
8 & \cmark &  & \cmark & 0.63 & 0.62 & 0.48 \\
8 & \cmark & \cmark & \cmark & 0.63 & 0.68 & 0.61 \\
\midrule
12 & \cmark &  &  & 0.64 & 0.61 & 0.44 \\
12 & \cmark & \cmark &  & 0.66 & 0.71 & 0.73 \\
12 & \cmark &  & \cmark & 0.61 & 0.63 & 0.48 \\
12 & \cmark & \cmark & \cmark & 0.63 & 0.69 & 0.61 \\
  \bottomrule
  \end{tabular}
  }
  \caption{Ablations on the Mistral on the three datasets. Here, $k$ is the number of demonstrations.}
  \label{tab:ablations-mistral}
  \end{table}

\section{Candidate set size can be smaller}
\label{appendix:candidate-set-size}

\begin{table}[!h]
\centering
\resizebox{\linewidth}{!}{
\fontsize{10pt}{12pt}\selectfont
\begin{tabular}{lccc}
  \toprule
  LLM & YouTube Slant & Ad Fontes News & AllSides News \\ 
  \midrule
  GPT-4o  & 0.66 & 0.74 & 0.52 \\
  Llama2  & 0.60 & 0.57 & 0.50 \\
  Mistral & 0.63 & 0.59 & 0.42 \\
  \bottomrule
  \end{tabular}
}
\caption{Accuracy of the three LLMs across all three datasets with candidate set size $|\mathcal{C}| = 500$ and $k=8$ demonstrations.}
\label{tab:smaller-candidate-set}
\end{table}

One might argue that the candidate set $\mathcal{C}$ from which demonstrations are selected is similar to the training set used for training supervised models and, thus, a large amount of annotated data is also required for LLMs.
While the size of the candidate set was configured high across all three datasets, we further analyze whether the need for annotated data can be alleviated by considering a smaller candidate set size as well.
Specifically, we set the candidate set size to only 500 samples, a fraction of what is generally required to train supervised models.
Testing this across all three LLMs and all three datasets in Table \ref{tab:smaller-candidate-set}, we observe that the accuracy is not majorly effected by this change.
This is likely because the \textsc{Set-BSR} approach already selects the most optimal demonstrations that maximize coverage.

\section{Balanced demonstration selection is better than random}
\label{appendix:balanced-vs-random}
We experiment with two demonstration selection techniques namely 1) \textbf{Random} where the demonstrations are randomly selected from the candidate set and 2) \textbf{Balanced BERTScore} where the BERTScore-Recall metric is used to select demonstrations in a label-balanced manner (equal liberal, neutral, and conservative demonstrations).
Comparing the performance of Mistral with 8 demonstrations, we see in Table \ref{tab:mistral-random-vs-balanced} that there is an average increase of $7\%$ in accuracy across the datasets when using Balanced BERTScore selection over random.

\begin{table}[h]
\centering
\resizebox{\linewidth}{!}{
\fontsize{10pt}{12pt}\selectfont
\begin{tabular}{lcc}
\toprule
Dataset        & Random & Balanced BERTScore\\
\midrule
YouTube Slant  & 0.52   & 0.61 \\
Ad Fontes News & 0.49   & 0.57 \\
AllSides News  & 0.39   & 0.42 \\
\bottomrule
\end{tabular}
}
\caption{Comparison of random demonstration selection and Balanced BERTScore demonstration selection using Mistral in a 8-shot setting. There is an average increase of $7\%$ in accuracy from random.}
\label{tab:mistral-random-vs-balanced}
\end{table}

\section{Qualitative results for the political news analysis for YouTube videos}
\label{appendix:qualitative-examples}

Qualitative results for political news videos, non-political news videos, and political non-news videos that changed predictions as a result of adding information about the source are provided in Tables \ref{tab:qual-polnews}, \ref{tab:qual-nonpolnews}, and \ref{tab:qual-polnonnews} respectively.

\begin{table*}[h]
\centering
\resizebox{\linewidth}{!}{
\begin{tabular}{ll}
\toprule
\textbf{Title} & \textbf{Channel} \\
\midrule

\multicolumn{2}{c}{\textbf{Liberal to Neutral}} \\
How welcome are Russian deserters in Georgia? | Focus on Europe & DW News \\
Ukrainians ditch Russian traditions and celebrate Christmas with the West & Sky News \\
Asian Americans Debate Model Minority \& Asian Hate | VICE Debates & VICE \\
Secret Service under scrutiny ahead January 6 committee's final report & PBS NewsHour \\
White House Correspondent Jeff Mason on his tense exchanges with Trump & Reuters \\
& \\
\midrule
\multicolumn{2}{c}{\textbf{Liberal to Conservative}} \\
Democrats introduce bill to block Trump from holding office & Newsmax \\
LIVE: Jan. 6 Committee Votes on Criminal Referrals, Final Report & NTD \\
New CRIMINAL REFERRALS Against Former President Trump Unveiled by Jan. 6th Committee & BlazeTV \\
Tonight's Tipping Points: Hate Crimes, Planned Parenthood, \& Trans Military Members! & One America News Network \\
Trolling \#AMFEST2022 Attendees as Left-Winger 'Garth Griffin' of Edge News & Project Veritas \\
& \\
\midrule
\multicolumn{2}{c}{\textbf{Neutral to Liberal}} \\
Haberman says Trump has been diminished but isn't dead politically & CNN \\
Kevin McCarthy's Path To Speaker Is In Peril & MSNBC \\
Anger grows in Russia over Ukraine's New Year's Day strike on Russian troops & CNN \\
Rep. Adam Kinzinger Delivers Final Speech As Congressman & MSNBC \\
'Shocked': Russia expert explains why Putin's change of word is significant & CNN \\
& \\
\midrule
\multicolumn{2}{c}{\textbf{Neutral to Conservative}} \\
UK Weather: What's behind Britain's big freeze? | UK snow chaos explained & Daily Mail \\
Transgender Clinical Psychologist on people de-transitioning | Dr. Erica Anderson & Newsmax \\
Actor slams crypto as 'biggest Ponzi scheme ever' during Senate hearing & Fox News \\
Rep. Spartz explains her reasoning for voting 'present' for Speaker & Fox News \\
Internet users allege Bryan Kohberger attended Idaho victims' vigil & Fox News \\
& \\
\midrule
\multicolumn{2}{c}{\textbf{Conservative to Liberal}} \\
Trump Supporter: Too Much Diversity Will Kill America & The Young Turks \\
Russian foreign minister: Ukraine must give up occupied territories & CNN \\
These Trump Supporters Are CONVINCED "Sleepy Joe Poopy Pants Isn't The Acting President" & The Young Turks \\
Steve Bannon reacts to Trump's 'major announcement' & CNN \\
Trump adviser tells reporter 'the magic is gone' in new article about re-election bid & CNN \\
& \\
\midrule
\multicolumn{2}{c}{\textbf{Conservative to Neutral}} \\
Justin Amash APPLAUDS Revolt Against McCarthy, Says Ruling Oligarchy CRUSHES Dissent & The Hill \\
State Of Emergency Declared At Texas Border & NBC News \\
Watch Trump claim the army retook airports during the Revolutionary War in July Fourth speech & The Globe and Mail \\
Trump Calls Jan. 6 Committee A 'Kangaroo Court' After Criminal Referrals & NBC News \\
'They shot up the WRONG house!' Sheriff Grady Judd on young kids shot at while in bed in Polk County & WFLA News Channel 8 \\

\bottomrule
\end{tabular}
}
\caption{Sample of political videos from news channels on YouTube whose predicted ideology changed as a result of adding information about the source.}
\label{tab:qual-polnews}
\end{table*}

\begin{table*}[h]
\centering
\resizebox{\linewidth}{!}{
\begin{tabular}{ll}
\toprule
\textbf{Title} & \textbf{Channel} \\
\midrule

\multicolumn{2}{c}{\textbf{Liberal to Neutral}} \\
TRIPLEDEMIC Or Free Market Failure To Blame For Medicine SHORTAGE? STRUGGLE To Find Children's Meds & The Hill \\
Watch: Jan. 6 Committee Presents Key Video Evidence Compilation & NBC News \\
& \\
\midrule
\multicolumn{2}{c}{\textbf{Neutral to Liberal}} \\
Top U.S. \& World Headlines — December 27, 2022 & Democracy Now! \\
Top U.S. \& World Headlines — January 3, 2023 & Democracy Now! \\
Raskin On Cancer Diagnosis: 'Totally My Plan To Make It Through This Thing' & MSNBC \\
Watch The 11th Hour Highlights: Dec. 20 & MSNBC \\
Plans For Zelenskyy D.C. Trip Came Together In A 'Matter Of Days' & MSNBC \\
& \\
\midrule
\multicolumn{2}{c}{\textbf{Neutral to Conservative}} \\
McCarthy speakership battle: What happens if he concedes? & Fox Business \\
The yule log Christmas tradition & Fox News \\
“Woman” Is Named Word of the Year & DailyWire+ \\
What is the controversy surrounding Meghan Markle and her sister? & Fox News \\
'DOUBLE WHAMMY': Real estate expert issues pricing warning & Fox Business \\
& \\
\midrule
\multicolumn{2}{c}{\textbf{Conservative to Liberal}} \\
All Gestures, All Shock Opera: DeSantis Launches His Newest Battle & MSNBC \\
Christian Influencer Makes Surprising Confession About Trans Porn & The Young Turks \\
The War on Christmas RAGES ON! & The Young Turks \\
Alex Jones Says He's Paying His Personal Trainer \$100,000 A Week & The Young Turks \\
'Look at where he is today': Bash on Kinzinger's farewell speech & CNN \\
& \\
\midrule
\multicolumn{2}{c}{\textbf{Conservative to Neutral}} \\
Cancelling the F-22 Fighter is Worse than You Think & Task \& Purpose \\

\bottomrule
\end{tabular}
}
\caption{Sample of non-political videos from news channels on YouTube whose predicted ideology changed as a result of adding information about the source.}
\label{tab:qual-nonpolnews}
\end{table*}

\begin{table*}[h]
\centering
\resizebox{\linewidth}{!}{
\begin{tabular}{ll}
\toprule
\textbf{Title} & \textbf{Channel} \\
\midrule

\multicolumn{2}{c}{\textbf{Liberal to Neutral}} \\
Is capitalism actually broken? & TED-Ed \\
Update from Ukraine | Ukraine is getting ready for the winter Attack | Ruzzia will lose it & Denys Davydov \\
Van Lathan: A Jewish Producer Confronted Kanye at TMZ when Ye Said He Loves Hitler (Part 2) & djvlad \\
The Trouble with the Electoral College & CGP Grey \\
Those Times Sam Harris Made Us Slap Our Knees & Enlightainment \\
& \\
\midrule
\multicolumn{2}{c}{\textbf{Liberal to Conservative}} \\
The New Beginning of Same-Sex Marriage & Ben Shapiro \\
Meet Kamala Harris's 6-Year-Old Speechwriter & The Babylon Bee \\
Pro-Choice Bosses - A Generous Offer To Our Pregnant Employees & Live Action \\
Confronting Charlie Kirk and Tim Pool at Turning Point USA America Fest & Alex Stein \\
I Showed Up To An Anti-Gun Protest In Los Angeles | Let's Talk & James Klug \\
& \\
\midrule
\multicolumn{2}{c}{\textbf{Neutral to Liberal}} \\
Meanwhile… Pantone's Color of the Year | Macron Makes Condoms Free In France & The Late Show with Stephen Colbert \\
Grandma Arrested For Feeding The Homeless?! Danbury Court Case Clarifcation + GIVEAWAY! & Long Island Audit \\
LIVE: EMERGENCY January 6 Committee hearing - Day 6 & Brian Tyler Cohen \\
The Funniest Donald Trump Stories On The Graham Norton Show | Part One & The Graham Norton Show \\
Prison Officer BUSTED After Massive Nazi Gang Attack Caught & The Damage Report \\
& \\
\midrule
\multicolumn{2}{c}{\textbf{Neutral to Conservative}} \\
Winning With The Hand You're Dealt with Rick Warren & Saddleback Church \\
Fighting the Darkest Criminal Enterprise on the Planet & The 700 Club \\
STORE WARS:  Target, Walmart \& Home Depot go to war against Organized Retail theft & News For Reasonable People \\
Warner MAY BE SUED over Amber Heard and Aquaman 2?! NEW Hollywood Lawsuit! & ThatUmbrellaGuy \\
Historical alteration on the Ukraine Map: Russian 1st Corps withdrawing en masse from 4 locations! & Divine Justice \\
& \\
\midrule
\multicolumn{2}{c}{\textbf{Conservative to Liberal}} \\
Elon Musk Finally CRACKS In Ultimate Outburst Over Free Speech & The Damage Report \\
Chris Wallace: This Would Never Have Happened Under Nancy Pelosi & The Late Show with Stephen Colbert \\
Mike Lindell Snaps After Hannity Admits There Was No Fraud & Tommy Campbell \\
Trump ATTACKS the FBI and DOJ and Loses his Mind in NEW POSTS & MeidasTouch \\
BREAKING: Kari Lake slammed with brutal legal news & Brian Tyler Cohen \\
& \\
\midrule
\multicolumn{2}{c}{\textbf{Conservative to Neutral}} \\
Why US Let Russia Become the Biggest Nuclear Power & The Infographics Show \\
The CIA's Biggest Scam Exposed & KiraTV \\
Cop Pulls Over "Legal Expert" And SCHOOLS HIM On The Law & Audit the Audit \\
Why US Enemies are Scared of Ghosts (Elite Special Forces) & The Infographics Show \\
"INSPIRATIONAL HOLIDAY VIDEO" — A Bad Lip Reading of Joe Biden & Bad Lip Reading \\
\bottomrule
\end{tabular}
}
\caption{Sample of political videos from non-news channels on YouTube whose predicted ideology changed as a result of adding information about the source.}
\label{tab:qual-polnonnews}
\end{table*}

\section{Placement of popular news channels based on the slant cut-offs}
\label{appendix:slant-cutoffs}

\begin{figure*}[h]
  \includegraphics[width=\textwidth]{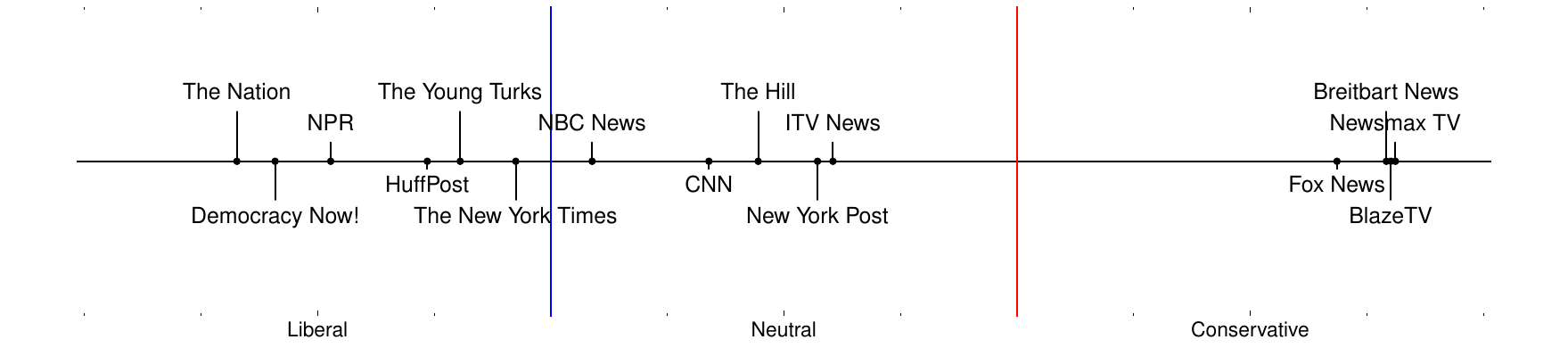}
  \caption{Placement of channels based on the slant cut-offs for the YouTube dataset.}\label{cutoffs}
\end{figure*}

Figure \ref{cutoffs} shows the placement of popular news channels based on the YouTube slant dataset. Corresponding cut-offs at -0.33 and +0.33 are also shown for the liberal and conservative respectively.

\section{Complete set of results for the various LLM configurations}
\label{appendix:full_results}

Table \ref{tab:final-results} shows the final set of results for all configurations.

\begin{table*}
\centering
\smallskip\noindent
\resizebox{0.85\linewidth}{!}{
\begin{tabular}{lccccccc}
 &  & \multicolumn{3}{c}{Metadata} & \multicolumn{3}{c}{Dataset} \\
\cmidrule(lr){3-5}  \cmidrule(lr){6-8}
Model & Shots & Title & Source & Body & YouTube Dataset & Ad Fontes News & AllSides News \\
\midrule
MLP & --- & \cmark & & & 0.49 & 0.38 & 0.41 \\
RoBERTa & --- & \cmark & & & 0.65 & 0.33 & 0.57 \\
\citet{baly2020detect} & --- & \cmark & & & --- & --- & 0.46 \\

\midrule
GPT-4o & 0 & \cmark &  &  & 0.64 & 0.69 & 0.50 \\
Llama2-13B & 0 & \cmark &  &  & 0.44 & 0.46 & 0.40 \\
Mistral-7B & 0 & \cmark &  &  & 0.59 & 0.55 & 0.44 \\
\midrule
GPT-4o & 0 & \cmark & \cmark &  & 0.52 & 0.78 & 0.74 \\
Llama2-13B & 0 & \cmark & \cmark &  & 0.43 & 0.54 & 0.64 \\
Mistral-7B & 0 & \cmark & \cmark &  & 0.67 & 0.72 & 0.71 \\
\midrule
GPT-4o & 0 & \cmark &  & \cmark & 0.57 & 0.75 & 0.55 \\
Llama2-13B & 0 & \cmark &  & \cmark & 0.49 & 0.55 & 0.46 \\
Mistral-7B & 0 & \cmark &  & \cmark & 0.59 & 0.62 & 0.41 \\
\midrule
GPT-4o & 0 & \cmark & \cmark & \cmark & 0.71 & 0.86 & 0.72 \\
Llama2-13B & 0 & \cmark & \cmark & \cmark & 0.47 & 0.67 & 0.58 \\
Mistral-7B & 0 & \cmark & \cmark & \cmark & 0.63 & 0.74 & 0.58 \\
\midrule
GPT-4o & 4 & \cmark &  &  & 0.67 & 0.74 & 0.55 \\
Llama2-13B & 4 & \cmark &  &  & 0.64 & 0.60 & 0.54 \\
Mistral-7B & 4 & \cmark &  &  & 0.61 & 0.57 & 0.42 \\
\midrule
GPT-4o & 4 & \cmark & \cmark &  & 0.73 & 0.87 & 0.83 \\
Llama2-13B & 4 & \cmark & \cmark &  & 0.68 & 0.70 & 0.75 \\
Mistral-7B & 4 & \cmark & \cmark &  & 0.65 & 0.73 & 0.73 \\
\midrule
GPT-4o & 4 & \cmark &  & \cmark & 0.69 & 0.80 & 0.58 \\
Llama2-13B & 4 & \cmark &  & \cmark & 0.63 & 0.68 & 0.51 \\
Mistral-7B & 4 & \cmark &  & \cmark & 0.60 & 0.61 & 0.43 \\
\midrule
GPT-4o & 4 & \cmark & \cmark & \cmark & 0.72 & 0.86 & 0.82 \\
Llama2-13B & 4 & \cmark & \cmark & \cmark & 0.58 & 0.77 & 0.68 \\
Mistral-7B & 4 & \cmark & \cmark & \cmark & 0.60 & 0.68 & 0.59 \\
\midrule
GPT-4o & 8 & \cmark &  &  & 0.68 & 0.73 & 0.55 \\
Llama2-13B & 8 & \cmark &  &  & 0.61 & 0.56 & 0.51 \\
Mistral-7B & 8 & \cmark &  &  & 0.62 & 0.60 & 0.44 \\
\midrule
GPT-4o & 8 & \cmark & \cmark &  & 0.72 & 0.87 & 0.86 \\
Llama2-13B & 8 & \cmark & \cmark &  & 0.62 & 0.76 & 0.74 \\
Mistral-7B & 8 & \cmark & \cmark &  & 0.64 & 0.71 & 0.73 \\
\midrule
GPT-4o & 8 & \cmark &  & \cmark & 0.70 & 0.81 & 0.59 \\
Llama2-13B & 8 & \cmark &  & \cmark & 0.57 & 0.67 & 0.52 \\
Mistral-7B & 8 & \cmark &  & \cmark & 0.57 & 0.61 & 0.47 \\
\midrule
GPT-4o & 8 & \cmark & \cmark & \cmark & 0.71 & 0.86 & 0.84 \\
Llama2-13B & 8 & \cmark & \cmark & \cmark & 0.47 & 0.78 & 0.67 \\
Mistral-7B & 8 & \cmark & \cmark & \cmark & 0.54 & 0.67 & 0.59 \\
\midrule
GPT-4o & 12 & \cmark &  &  & 0.69 & 0.74 & 0.55 \\
Llama2-13B & 12 & \cmark &  &  & 0.60 & 0.57 & 0.50 \\
Mistral-7B & 12 & \cmark &  &  & 0.64 & 0.61 & 0.44 \\
\midrule
GPT-4o & 12 & \cmark & \cmark &  & 0.74 & 0.87 & 0.87 \\
Llama2-13B & 12 & \cmark & \cmark &  & 0.54 & 0.72 & 0.72 \\
Mistral-7B & 12 & \cmark & \cmark &  & 0.66 & 0.71 & 0.73 \\
\midrule
GPT-4o & 12 & \cmark &  & \cmark & 0.71 & 0.82 & 0.59 \\
Llama2-13B & 12 & \cmark &  & \cmark & 0.46 & 0.62 & 0.49 \\
Mistral-7B & 12 & \cmark &  & \cmark & 0.53 & 0.61 & 0.46 \\
\midrule
GPT-4o & 12 & \cmark & \cmark & \cmark & 0.71 & 0.87 & 0.84 \\
Llama2-13B & 12 & \cmark & \cmark & \cmark & 0.40 & 0.77 & 0.67 \\
Mistral-7B & 12 & \cmark & \cmark & \cmark & 0.51 & 0.67 & 0.59 \\
\midrule

\end{tabular}
}
\caption{Results showing the accuracy of all experiments with GPT-4o, Llama2, Mistral across all three datasets and number of shots. Shown at the top are baseline results including the results from \citet{baly2020detect}.}
\label{tab:final-results}
\end{table*}

\section{Code and Reproducibility}

Our code and data used for prompting the models and analyzing the results can be found here: \url{https://github.com/Haroon96/Ideology-Estimation-LLMs}

\end{document}